# TriggerNet: A Novel Explainable AI Framework for Red Palm Mite Detection and Multi-Model Comparison and Heuristic-Guided Annotation

[1]Harshini Suresha*, [2]Kavitha SH

[1,3,4]Department of Biotechnology, PES University, BSK III stage, Bengaluru - 560085;

Mail id: harshinisuresha7@gmail.com

**Abstract:** The red palm mite infestation has become a serious concern, particularly in regions with extensive palm cultivation, leading to reduced productivity and economic losses. Accurate and early identification of mite-infested plants is critical for effective management. The current study focuses on evaluating and comparing the ML model for classifying the affected plants and detecting the infestation. TriggerNet is a novel interpretable AI framework that integrates Grad-CAM, RISE, FullGrad, and TCAV to generate novel visual explanations for deep learning models in plant classification and disease detection. This study applies TriggerNet to address red palm mite (Raoiella indica) infestation, a major threat to palm cultivation and agricultural productivity. A diverse set of RGB images across 11 plant species, Arecanut, Date Palm, Bird of Paradise, Coconut Palm, Ginger, Citrus Tree, Palm Oil, Orchid, Banana Palm, Avocado Tree, and Cast Iron Plant was utilized for training and evaluation. Advanced deep learning models like CNN, EfficientNet, MobileNet, ViT, ResNet50, and InceptionV3, alongside machine learning classifiers such as Random Forest, SVM, and KNN, were employed for plant classification. For disease classification, all plants were categorized into four classes: Healthy, Yellow Spots, Reddish Bronzing, and Silk Webbing. Snorkel was used to efficiently label these disease classes by leveraging heuristic rules and patterns, reducing manual annotation time and improving dataset reliability.

**Author's Note:**
All figures and visualizations presented in this preprint were originally developed for our NeurIPS 2025 workshop paper titled *"A Multi-Method Interpretability Framework for Probing Cognitive Processing in Deep Neural Networks across Vision and Biomedical Domains."* That shorter, preliminary version was accepted at the NeurIPS 2025 Workshops on *SPiGM* and *Reliable ML from Unreliable Data*. The present manuscript extends that work by introducing the complete TriggerNet framework, expanded experiments, and heuristic-guided annotation strategy. All images are reused here with updates and extended explanations.

**Key words:** TriggerNet framework; Red palm mite; Snorkel; Grad-Cam; FullGrad; RISE; TCAV

## 1 Introduction

The red palm mite (RPM) also is a highly crop destructive pest known for infesting the undersides of palm leaves which can cause an extensive and widespread agricultural loss if left unchecked. This pest appears bright red in color with an oval shape and its job is to feed on the plant's cellular content resulting in yellowing of leaves, a phenomenon also called as chlorosis. It can also lead to other conditions like necrosis and defoliation. The former one is the death of the internal body tissues caused by violent uninhibited process that damages the cells and the latter one causes the tree to loosen all its leaves which significantly reduces photosynthesis and overall plant health. These host plants of the pest include ornamental and economically important species like Coconut, Areca nut, Date palm, Cast iron, and Bird of Paradise. It is mainly reported in regions of tropical and subtropical spanning across India and Sri Lanka in east, South America and Brazil in the west, southern regions like the Maldives and Florida in the north. The intensity of damage is influenced by the factors such mite population density, climatic circumstances and the vulnerability of the host plant. To identify and detect Red Palm Mite-affected palms and leaves, multiple machine learning and deep learning algorithms were employed. The classification model used CNN, EffecientNet, MobileNet, ViT, ResNet50, InceptionV3, RF, SVM and KNN classifier. For detection tasks, CNN and YOLOv8 models were used to accurately identify the presence of RPM symptoms in affected plants.

TriggerNet is a novel interpretability framework that integrates Grad-CAM, FullGrad, RISE, and TCAV to enhance the reliability of deep learning models used in plant health monitoring. By capturing spatial relevance (Grad-CAM, FullGrad), probabilistic feature importance (RISE), and concept-level reasoning (TCAV), TriggerNet localizes disease-specific symptoms like yellow spots, silk webbing, and bronzing while distinguishing them from background noise. Unlike conventional single-method explanations, this hybrid approach allows domain experts to validate model decisions based on physiologically meaningful plant traits. In practice, TriggerNet has exposed detection failures in ViT (e.g., missing subtle webbing) and inattentiveness in MobileNet (e.g., early-stage symptoms), directly guiding model refinement. When integrated with models like ViT, YOLOv8, and CNNs, TriggerNet not only boosts interpretability but also supports trustworthy AI deployment in real-world agricultural settings. What distinguishes TriggerNet from existing explainable AI approaches is its multi-model, multi-method fusion design.

## 2 Related Work

This research paper developed a CNN model (VGG-19 architecture) trained on PlantVillage to achieve a reported 95.6% accuracy in classifying healthy and diseased plant leaves [1]. The study checks if a newer computer learning model, EfficientNet, can outperform established models like AlexNet, ResNet50, VGG16, and Inception V3 when trained on the PlantVillage dataset where they concluded that EfficientNet achieved a greater accuracy as compared to others [2]. MobileNetV3-Large achieved 75.2% top-1 accuracy on ImageNet, while MobileNetV3-Small obtained 67.668% top-1 accuracy, showcasing improvements over MobileNetV2 with reduced latency. [3]. The effectiveness of a traditional machine learning approach, KNN algorithm, combined with GLCM feature extraction and K-means segmentation, for automated plant disease detection using leaf images from the Plant Village dataset, achieved a reported 93% accuracy[4]. The main goal was to automate plant leaf disease detection by using GLCM feature extraction, followed by classification using both SVM and RF classifiers utilizing a dataset of diverse plant leaf images exhibiting symptoms of various bacterial and fungal diseases to identify and classify diseases[5]. This research explored the use of YOLOv3 and YOLOv4 analyzing healthy and diseased peach and strawberry leaves where YOLOv4 outperformed YOLOv3 in less time[6]. The integration of Grad-CAM with ResNet152 improved the transparency of corn leaf disease diagnosis, providing interpretable heatmaps that align with expert assessments[7]. Integrating Grad-CAM with ViT improved clarity in detecting leaf rust disease, aligning model attention with symptomatic regions[8]. This study achieved high accuracy (mAP consistently in the 85-90%+ range) in plant disease detection by combining YOLOv5 for localization with Grad-CAM for Explainability[9].

## 3 Proposed system

In this study, we propose a novel dual-stage deep learning framework for early and accurate identification of Red Palm Mite infestation in tropical plants. The system comprises two principal stages, Plant and Disease Classification and Infestation Detection, with interpretability deeply embedded to promote transparency and trust. CNN, ResNet50, EfficientNet, ViT, MobileNet, InceptionV3, Random

Forest, SVM, and KNN were the nine distinct architectures used for plant classification. Yolov8 and CNN were the two algorithms for detection. Every algorithm is essential to enhancing classification precision and guaranteeing reliable performance. The classification system follows a structured pipeline divided into the following six major stages.

1) Dataset and Splitting Strategy: The dataset used in TriggerNet comprises 11,550 images of Red Palm Mite-affected and healthy plants, collected from publicly available sources like Kaggle, Roboflow, Mendeley data, Manipal, forestry images and field photography collected from our very own field[10,11]. A 90:10 train-test split was used for CNN, EfficientNet, MobileNet, ViT, ResNet50, and InceptionV3, while an 80:20 split was used for SVM, RF, and KNN. For disease classification, the dataset was labeled using Snorkel in CSV format which was used to create supervision pipelines using heuristic labeling based on color patterns, texture, and image metadata. It had 4 columns and 10,450 rows. The dataset is split into training, validation and testing sets in a 70:15:15 ratio, ensuring class balance. The CNN and YOLOv8 models were used to accurately identify the presence of RPM symptoms in affected plants.

2) Pre-processing of Sample Image: the dataset contained images in both RGB and grayscale formats. Since field images were captured in RGB format, grayscale images were converted to RGB for uniformity. ResNet50 images were resized to 224x224, EfficientNet to 132x132, and InceptionV3 to 299x299. To improve model generalization, data augmentation techniques were applied, including rotation (random rotations within a 20° range), flipping (horizontal and vertical), zooming (random zoom levels between 0.8x to 1.2x), and brightness adjustments (±20% variation). The RGB follows the standard mapping:

$$R=Igray,\ G=Igray,\ B=Igray \quad (1)$$

Auto-orientation was applied to standardize image angles and static cropping was performed on 25-75% horizontal and vertical regions and flipped where each training sample was augmented to generate three outputs, 90° rotations (clockwise, counter-clockwise, and upside down), and 12% maximum zoom. The shear transformations (±15° horizontally and vertically) and saturation adjustments (±25%) were applied to increase the model's robustness to image variations.

3) Segmentation of sample images: Segmentation was important for isolating the leaf area from the background. Watershed segmentation, leveraging markers for foreground and background regions, effectively identified leaf areas.

$$\sigma_b^2 = w_1(t) \cdot w_2(t) \cdot [\mu_1(t) - \mu_2(t)]^2 \quad (2)$$

Otsu's thresholding method was applied to maximize variance between foreground and background, determining an optimal threshold value for improved binarization. Otsu's thresholding formula calculated class probabilities and means to isolate significant image regions efficiently.

4) Thresholding: It helped improve contrast, especially for distinguishing healthy leaves from those showing early mite symptoms. Adaptive thresholding dynamically adjusted the threshold value across different image regions, enhancing detail preservation.

5) Feature Extraction: For CNN-based models like ResNet50, EfficientNet, ViT, MobileNet, and InceptionV3, convolutional layers automatically extracted critical features. Conversely, for machine learning models like Random Forest, SVM, and KNN, handcrafted features such as color histograms, texture features using GLCM, and edge detection filters like Sobel and Canny operators were employed to capture essential leaf characteristics.

$$P(i,j \mid d,\theta) \quad (3)$$

The orientation ranged from 0°, 45°, 90° to 135°. The GLCM formula for texture analysis is given in Eq. (3).

6.) Classification Algorithms: The first stage of the proposed system focuses on classifying the input plant image into one of 11 predefined species, including Areca Nut, Banana Palm, and others. Accurate species identification is crucial, as RPM symptoms manifest differently across host plants. This stage uses deep learning models trained on RGB images to extract visual features and classify plant types forming the basis for context specific disease detection.

7.) Symptom detection: After species identification,

the second stage detects the presence and severity of RPM infestation symptoms. Using a bounding box-based detection framework, the system localizes visual cues such as initial chlorotic speckling, progressive reddish discoloration due to chlorophyll degradation, mite webbing and dense fibrous webbing.

8.) TriggerNet Framework: The pipeline begins by feeding the preprocessed RGB plant image (resized to 224×224 for CNN and ViT, and 640×640 for YOLO) into three parallel branches. Each branch represents a distinct model, a CNN (ResNet50), a ViT (ViT-B/16), and a YOLOv8 variant. Instead of modifying the model architectures, TriggerNet leverages their frozen weights and taps into their intermediate representations to extract meaningful feature responses. Grad-CAM is applied to the final convolutional or attention-based layers (layer4 in CNN, [CLS] token attention in ViT, and final detection backbone in YOLO), highlighting class-specific spatial regions. RISE uses a set of N random binary masks (N=4000) and randomized input perturbations that aggregates model outputs to assign importance to image pixels. FullGrad captures input and bias gradients throughout the network to provide fine-grained attribution.

Once saliency maps are extracted from each method, they are spatially normalized and fused within each model stream using a weighted averaging scheme. This intra-model fusion results in a single interpretability map per model, combining Grad-CAM, RISE, and FullGrad in a unified representation give in Eq. (4). The fused saliency maps from the CNN, ViT, and YOLO models are then further aggregated through inter-model fusion, generating a final interpretability map that encapsulates local (CNN), global (ViT), and detection-based (YOLO) explanations given in Eq. (5).

$$S_{Trigger}=(S_{Grad\text{-}CAM}+S_{RISE}+S_{FullGrad}) \quad (4)$$

$$S_{Trigger}=(S_{CNN}+S_{ViT}+S_{YOLO}) \quad (5)$$

The final output of TriggerNet includes class predictions, bounding boxes (for YOLO-based detection), and a fused saliency overlay map that visually justifies the model's decision. The saliency outputs are quantitatively evaluated using interpretability-specific metrics such as the Pointing Game accuracy, mean Intersection over Union (mIoU), TCAV scores, and deletion/insertion AUC.

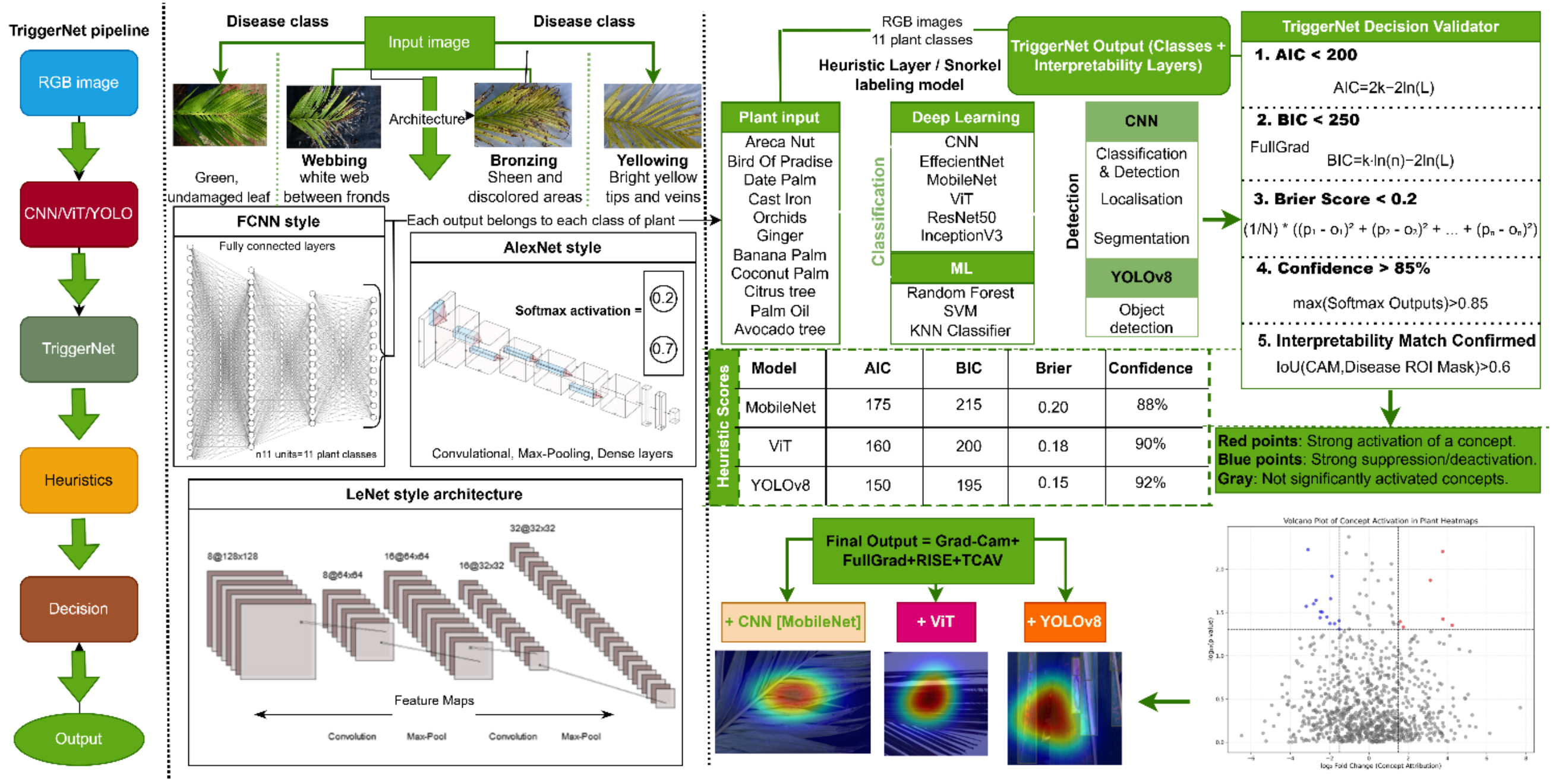

**Figure 1: TriggerNet Framework Integrating CNN, ViT, and YOLOv8 Architectures with Heuristic-Based Decision Validation for Plant Disease Classification and Detection**

# 4. Methodology

## 4.1 Dataset curation

All images were resized to 224×224 pixels and normalized to the (0,1) range. Data augmentation techniques including random horizontal flips, brightness adjustments, and slight rotations were applied to increase dataset diversity. Weak labels for disease severity were refined using Snorkel's labeling functions. After cleaning, the final dataset contained 3,800 unaffected, 4,200 mildly affected, and 2,450 severely affected images. Snorkel leveraged multiple labeling functions (LFs) to assign these labels based on predefined rules and visual characteristics. Each LF followed the form $\lambda i : X \rightarrow Y \cup \{\emptyset\}$, where each function either assigned a class label or abstained if uncertain. The outputs from multiple LFs were then aggregated using Snorkel's probabilistic model, which employed a weighted majority vote to determine the most likely label. This was calculated as:

$$\hat{y} = argmax \sum_{i=1}^{n} w_i \cdot \lambda i(x)$$

(6)

where, $w_i$ denotes the reliability weight of the $i^{th}$ LF. This ensured higher accuracy by prioritizing LFs with better performance. In our model there were 4 LFs:

a.) $\lambda_1$ detects Yellow Spots based on colour features (early chlorosis),

b.) $\lambda_2$ detects Silk Webbing using texture patterns (mite colonies, web structures),

c.) $\lambda_3$ detects Healthy Leaves by checking for no visible damage (no visible infestation),

d.) $\lambda_4$ detects reddish brown by checking the chlorophyll loss (advanced chlorosis).

## 4.2 Model Architectures

i.) Convolutional Neural Network (CNN)

CNN consisted of multiple convolutional layers followed by pooling layers to extract spatial hierarchies. The convolutional layers applied kernels that performed element-wise multiplication with input pixel values, followed by summation to generate feature maps. The ReLU activation function introduced non-linearity of f(x) = max(0,x). The categorical cross-entropy loss function optimized model performance using the Adam optimizer. Dropout layers were integrated to reduce overfitting, and the final dense layer employed softmax activation for multi-class classification. It utilized Eq. (7) for this approach.

$$P(y=i|x) = \frac{e_i^z}{\sum_{j=1}^{C} e_j^z} \quad (7)$$

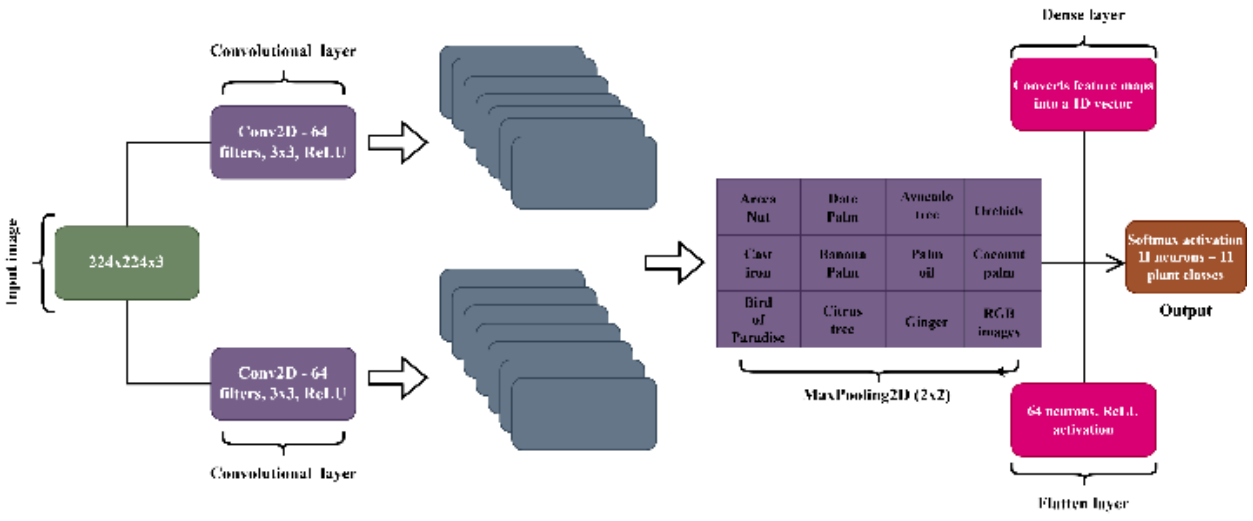

**Figure 2: CNN-based plant classifier pipeline**

ii.) ResNet50

ResNet50 utilized residual connections that bypassed one or more layers solving vanishing gradient issues. Since RPM symptoms (like yellow spots or silk webbing) involve subtle texture changes, deeper networks can easily lose these fine details. ResNet's skip connections preserved those patterns across layers. Its architecture included convolutional blocks with batch normalization and ReLU activation. Residual blocks ensured that both low-level (edge details) and high-level (complex mite damage patterns) features were combined. The identity mapping layer ensured gradient flow by adding input features directly to deeper layers of y = F(x) + x. With 50 layers, ResNet50 had the depth required to analyze detailed patterns like leaf discoloration, webbing structures, and mite feeding marks, which are essential indicators of infestation. Field images often include background clutter such as soil, plant debris, and uneven lighting. ResNet50's architecture effectively filtered out this noise while isolating disease features that helped in improving the convergence.

iii.) EffecientNet

We employed compound scaling to uniformly scale network depth, width, and resolution. The model's core included inverted residual blocks with depth-wise convolutions, optimizing both efficiency and accuracy. Inverted Residual blocks with linear bottlenecks captured detailed patterns while maintaining lightweight architecture. EfficientNet's ability to compress information into bottleneck layers reduced overfitting, especially given the 90:10 data split in CNN models. Squeeze-and-excitation mechanisms recalibrated channel-wise feature maps to improve feature representation emphasizing disease-specific patterns like reddish bronzing and silk webbing while suppressing irrelevant background details. EfficientNet effectively adapted to diverse leaf structures (e.g., broad coconut leaves, slender Arecanut leaves) by efficiently learning spatial hierarchies.

iv.) Vision Transformer (ViT)

The self-attention mechanisms to capture long-range dependencies across image patches was applied here. Input images were divided into fixed-size patches and mite symptoms across plant species. Despite its reduced complexity, MobileNet effectively detected clearer symptoms, such as fully damaged leaves with distinct discoloration or webbing. However, it struggled with early-stage symptoms that appeared as subtle visual changes.

vi.) InceptionV3

Asymmetric convolutions and factorized filters to reduce parameter count was used here. Multiple filter sizes were applied in parallel convolutional paths to capture diverse feature scales effectively. Auxiliary classifiers improved gradient propagation, enhancing convergence. InceptionV3's core strength lies in its Inception modules, which apply 1x1, 3x3, and 5x5 convolution filters in parallel. This multi-scale design allows the network to detect features of different sizes. For RPM detection, this worked well for identifying bronzing (which appears as widespread discoloration) and webbing patterns (which span larger portions of the leaf). While InceptionV3's multi-scale filtering is powerful, it may still struggle with extremely small or subtle features like tiny yellow spots in the early stages of mite infestation. Such details may get diluted when filters operate on larger receptive fields.

vii.) Random forest

projected into linear embeddings. The transformer encoder used multiple self-attention heads, which computed weighted averages of input features, here Q, K, V is the query, key and value matrices.

$$Attention(Q,K,V) = softmax\,)\,V \quad (8)$$

v.) MobileNet

Depth-wise separable convolution was used to separate spatial and channel wise convolutions. The concept in Eq. (8) was used for this architecture.

$$\text{DepthwiseConv}(x)\cdot\text{PointwiseConv}(x) \quad (9)$$

It had a considerable difference in training and testing accuracy that is reported in the results section. The reason is because unlike deeper AI networks like ResNet50, MobileNet's streamlined architecture has fewer layers, which can limit its ability to extract intricate patterns required for distinguishing similar

Random Forest was employed as one of the ML classifiers to distinguish between healthy and affected plant leaves. It doesn't automatically extract features like CNNs, handcrafted features were extracted during pre-processing and feature engineering. This approach combined color, texture, and structural information to analyze the dataset effectively. RGB histograms captured pixel intensity distributions across red, green, and blue channels to identify color differences like yellow spots, bronzing, or healthy green regions. Color moments (mean, standard deviation, and skewness) moments quantified variations in color intensity, which are prominent when distinguishing mite-induced discoloration. Laplacian and Sobel edge detectors extracted sharp edge transitions that are useful in identifying webbed regions. The extracted features were then used to train a RF model with 100 number of trees, and maximum depth of 15. GLCM captured textural patterns like roughness, contrast, and uniformity in leaf surfaces especially effective for detecting webbing or damaged leaf veins. The contrast metric in GLCM was calculated as:

$$Contrast = \sum_{i=0}^{N-1}\sum_{j=0}^{N-1}(i-j)^2\cdot P(i,j) \quad (10)$$

Sobel Filter and Canny Edge Detection extracted sharp transitions and boundary details, highlighting RPM

webbing and leaf edge distortions. RF created multiple subsets from the original dataset, where each subset $D_b$ was sampled with replacement. Each subset retained the same number of samples as the original dataset but included duplicated entries of $D_b = \{X_{b1}, X_{b2},.., X_{bn}\}$. This randomness ensured each tree explored different feature combinations, improving generalization. We also used Gini Index to determine the optimal split at each node. For a dataset containing both healthy and affected leaf samples, Gini impurity was calculated as:

$$Gini = 1 - \sum_{i=1}^{C} p_i^2 \quad (11)$$

Here, C is the number of classes and in our model's case there are 4 classes that is Healthy, Yellow spots, Reddish Bronzing and Silk webbing. If a node had 50% healthy, 30% yellow spots and 20% bronzing then,

$$Gini=1-(0.5^2+0.3^2+0.2^2)=1-(0.25+0.09+0.04) =0.6$$

In another case if a node contained 6-% healthy, 30% yellow spots and 10% bronzing then,

$$Gini=1-(0.6^2+0.3^2+0.1^2)=1-(0.36+0.09+0.01)=0.54$$

The main goal of using Gini impurity was to ensure that the child nodes had a higher concentration of samples belonging to a simple class and secondly to verify that there was no class overfitting. Then each decision tree independently predicted the class for an image. The final prediction used majority voting that is $\hat{y} = mode\{h_1(x), h_2(x),...,h_B(x)\}$. Any missing or under-represented patterns could lead to misclassification. Adding Fourier transforms (to capture periodic patterns) along with Haralick texture features (to extract second order statistical texture) and HoG (Histogram of Oriented Gradients that focused on capturing edge structures and gradients) later enhanced feature depth.

ix.) SVM (Support Vector Machine)

We used the kernel trick to map data into higher dimensional space further optimizing the hyperplane separation. The RBF kernel improved boundary flexibility to handle complex feature distributions.

$$K(X_i, X_j) = exp(-\gamma \, || X_i - X_j ||^2) \quad (12)$$

To enhance SVM's capability, additional techniques such as Wavelet Transform, LBP (Local Binary Pattern), and Zernike Moments can significantly improve the feature representation. Wavelet Transform effectively captures both spatial and frequency information, making it highly suitable for detecting Red Palm Mite-induced symptoms like bronzing, webbing, and structural distortions. Unlike Fourier Transform, Wavelets provide localized frequency details, which are critical for identifying fine-grain leaf damage patterns.

$$W(a, b) = \int f(t) \cdot \psi \cdot (t/a - b) \, dt \quad (13)$$

We also applied Local Binary Pattern because it encodes pixel intensity differences to identify patterns in the leaf surface. This is highly effective in detecting mite webbing's fine-textured regions or bronzed leaf surfaces. Zernike Moments enhanced the model's understanding of the leaf shape deformation by isolating structural differences such as distorted veins and leaf margins increasing the model's accuracy.

x.) KNN (k-Nearest Neighbors)

This ML algorithm helped in classifying data points by calculating distances to its nearest neighbors using the Euclidean distance metrics. Color histograms effectively captured distinct color changes like bronzing, webbing, and chlorosis caused by mite infestations. By analyzing pixel intensity distributions in RGB channels, the KNN model leveraged these color variations to improve classification performance. Color Histogram Probability Distribution Equation was equated in our model as:

$$p_k = \frac{h_k}{N} \quad (14)$$

xi.) YOLOv8 (You only look once)

This version of YOLO was used as this architecture was particularly suited for this task due to its efficient single-shot detection mechanism, which processes the entire image in one forward pass, making it ideal for fast and accurate detection. The input images were resized to 640x640 dimensions to match YOLOv8's optimal input size, ensuring enhanced precision during the detection process. The YOLOv8 model utilized CSPDarknet53 as its backbone, which effectively captured both low-level visual details and high-level semantic features. This structure improved the model's ability to detect RPM symptoms, particularly webbing and bronzing, even in complex field environments. The PANet (Path Aggregation Network) served as the neck structure, merging spatially rich low-level features with deeper abstract features to improve localization accuracy. YOLOv8's detection head predicted bounding box coordinates, class probabilities, and confidence scores. Bounding box dimensions were calculated using the following equations:

$$\hat{X} = \sigma(t_x) + c_x \quad (15)$$

$$\sigma(t_y) + c_y \quad (16)$$

$$\hat{w} = p_w \cdot e^t{}_w \quad (17)$$

$$\hat{h} = p_h \cdot e^t{}_h \quad (18)$$

To refine the predictions, Non-Maximum Suppression (NMS) was applied in the post-processing stage. NMS eliminated overlapping boxes by retaining only the highest-confidence detections, ensuring precise localization of infected regions. This step was vital in identifying multiple symptoms across different parts of the same plant.

**4.3 TriggerNet Interpretability Stack**

TriggerNet adopts a Hierarchical Interpretability Stack (HIS) to dynamically select the most suitable interpretability technique based on the model type (CNN, YOLO, ViT) and output uncertainty. For transformer-based architectures like ViT, the stack prioritizes FullGrad + Grad-CAM, as they provide superior token-level attributions by leveraging both spatial relevance and gradient propagation through attention maps. For convolutional networks such as CNNs, ResNet's, and Inception modules, TriggerNet activates combinations like Grad-CAM + TCAV or RISE + Grad-CAM, which offer reliable spatial heatmaps and concept traceability. The stack selection is selected by an internal controller that gives a score:

$$S_{int} = \arg\max_{i\in\{1,2,3\}} (\lambda_i \cdot I_i + \gamma \cdot Uncertainity) \quad (19)$$

where, $I_i$ denotes interpretability confidence from method $I$, $\lambda_i$ is the method weight (learned via training-time AUC gain), γ denotes uncertainty amplification factor from Softmax entropy.

The Model-Aware Interpreter Assignment (MAIA) system pairs each classification or detection model with its optimal interpretability techniques using a learned graph-based meta-model. In this graph structure, the Nodes represent models (e.g., ViT, YOLOv8) and interpretability methods (e.g., TCAV, RISE). Edges are scored based on three key compatibility metrics that is Locality Fidelity (how spatially precise the method is), Concept Traceability (how well the method links features to known concepts), Perturbation Robustness (how stable the method is under input perturbations). The final assignment is computed via the formula:

$$Assign_{model} = \arg\max_{method} (\alpha \cdot LF + \beta \cdot CT + \delta \cdot PR) \quad (20)$$

The hyperparameters α,β,δ are tuned to maximize validation interpretability fidelity.

To unify multiple interpretability signals, TriggerNet incorporates an Interpretability Fusion Module (IFM). This module combines saliency maps or concept scores across various interpretability techniques using layer-wise attention gating. First, all maps are normalized to the range [0, 1] and resized using bilinear interpolation to ensure spatial alignment. Then, each map $M_i$ from method $i$ is weighed with a learned attention mask producing the fused saliency map:

$$M_{fused} = \sum_{i=1}^{N} A_i \odot M_i \quad (21)$$

Here, ⊙ denotes element-wise multiplication. The resulting fused map is not only class-specific and model-aware, but also concept-validated, ensuring consistency and robustness in interpretability outputs.

TriggerNet does not apply interpretability uniformly across all samples. Instead, it employs a Trigger Decision Mechanism (TDM) that activates interpretability only when prediction confidence or label quality is questionable. Specifically, interpretability is triggered if:

a.) The prediction entropy exceeds 0.3,

b.) The ensemble agreement falls below 0.75,

c.) The prediction lies near class boundaries in t-SNE space,

d.) If the disease category was weakly labeled (e.g., via Snorkel), and requires visual validation.

This selective triggering ensures efficient and targeted explanation, focusing interpretability only where it's most needed. After extracting concept importance using TCAV, TriggerNet employs a Concept Alignment Layer (CAL) to validate whether these conceptual insights align with the spatial heatmaps from Grad-CAM. This is done using cosine similarity between the two attribution vectors:

$$AlignmentScore = cos(\vec{\vartheta}_{TCAV}, \vec{\vartheta}_{Grad\text{-}Cam}) \quad (22)$$

Only those concept activations with an alignment score >0.6 are retained filtering out noisy concepts and maintaining only those with a strong correspondence to the model's spatial focus. TriggerNet supports iterative improvement by comparing interpretability feedback to misclassification zones. For example, if a Grad-CAM map consistently activates on background (not leaf), those samples are flagged. During training, TriggerNet introduces a Saliency-Concept Consistency Loss (SC² Loss) to align spatial and concept-level explanations. It encourages coherence between Grad-CAM saliency maps and TCAV concept masks using Intersection over Union (IoU):

$$L_{SC^2} = 1 - IoU\,(M_{Grad\text{-}Cam}, M_{TCAV\text{-}Concept}) \quad (23)$$

Minimizing this loss during fine-tuning reinforces internal consistency, ensuring that the model's spatial attention aligns with high-level concepts.

### 4.4 Interpretability Techniques Used in TriggerNet

#### 4.4.1 Grad-Cam

Grad-CAM is one of the core interpretability techniques integrated into TriggerNet for convolution-based architectures like CNNs, ResNet, and YOLO. It provides spatially meaningful visual explanations by producing class-discriminative heatmaps, highlighting regions in the input image that contribute most to a model's decision. This is particularly useful for understanding model predictions on plant images where disease features (e.g., discoloration, texture changes) may appear in localized regions.

The standard Grad-CAM formulation by Selvaraju et al.[12] involved generating heatmaps by computing the gradient of the class score $y^c$ w.r.t to the activation maps $A^k$ from the last convolutional layer $\partial_{y^c}/\partial_{A^k}$. The gradients are then globally averaged to obtain the importance weight for each channel $k$ and the final saliency map is obtained by a weighted combination of feature maps followed by a ReLU operation:

$$\alpha_k^c = \frac{1}{Z} \sum_i \sum_j \frac{\partial_{y^c}}{\partial A_{ij}^k}, \; Z = H \times W \quad (24)$$

$$L_{GradCam}^c = ReLU \sum_k \alpha_k^c A^k \quad (25)$$

This heatmap is then resized to match the input dimensions and is overlaid on the original image to interpret which regions influenced the class prediction.

In TriggerNet, the Grad-CAM method is customized to support both classification and detection tasks for both CNN and YOLO models to generate precise attribution maps within the classification and detection workflows. Given an input image $I \in R^{H \times W \times 3}$, feature maps are extracted from a specific convolutional layer L as $\Phi^L(I)$. The notation is adapted to remain consistent with the rest of the TriggerNet pipeline, and post-processing (normalization + upsampling) is applied explicitly for fusion and loss alignment purposes. The gradient of the class score $Sc$ w.r.t maps is calculated as:

$$G_c^L = \frac{\partial S_c}{\partial \Phi^L(I)} \quad (26)$$

The channel-wise importance weight is given by:

$$\delta_k^c = \frac{1}{h \cdot w} \sum_{i=1}^{h} \sum_{j=1}^{w} G_c^L[i,j,k] \quad (27)$$

Using these weights, the class-specific saliency map is derived as:

$$\Gamma_{GradCam}^c = ReLU\,\Big(\sum_{k=1}^{d} \delta_k^c \cdot \Phi_{[:,:,k]}^L\Big) \quad (28)$$

$$\Gamma_c^* = Upsample\,(\Gamma_{GradCam}^c, H, W) \quad (29)$$

In YOLO-based detection tasks, this process is applied to the last convolutional feature map just before the detection head, ensuring heatmaps align with object regions (e.g., bounding boxes of diseased leaves). In transformer-based models like ViT, Grad-CAM is substituted with token-based attribution, which is handled in later sections under FullGrad and TCAV.

#### 4.4.2 FullGrad

FullGrad extends traditional attribution methods by capturing input-level gradients and bias/intermediate contributions from every layer of the model, rather than just relying on the final convolutional layer as in Grad-CAM. It computes FullGrad saliency maps across, Transformer MLP biases in ViT, Residual bias paths in YOLO heads, and Attention-weighted intermediate blocks in detection branches. Building on the original FullGrad formulation of Srinivas et al[13]., we adapt it to the TriggerNet architecture to integrate both input-level and intermediate-layer attributions. The attribution is computed in 3 steps:

a.) Input-level Gradient Term:

$$G_x = x \odot \frac{\partial f(x)}{\partial x}$$

(30)

b.) Bias Gradient Contribution from All Layers:

$$G_b^l = b^l \odot \frac{\partial f(x)}{\partial b^l}$$

(31)

c.) Total FullGrad Attribution:

$$\text{FullGrad}(x) = G_x + \sum_l G_b^l \quad (32)$$

d.) Normalize & Rescale:

$$\Gamma_c^* = NormUpsample\ (\Gamma_{GradCam}^c, H, W$$

(33)

Heatmaps from FullGrad showcasing layer-wise bias and gradient contributions (features 16, 23, 30) for CNN, ViT, and YOLOv8 models on disease-affected plant inputs is visualised in figure3.

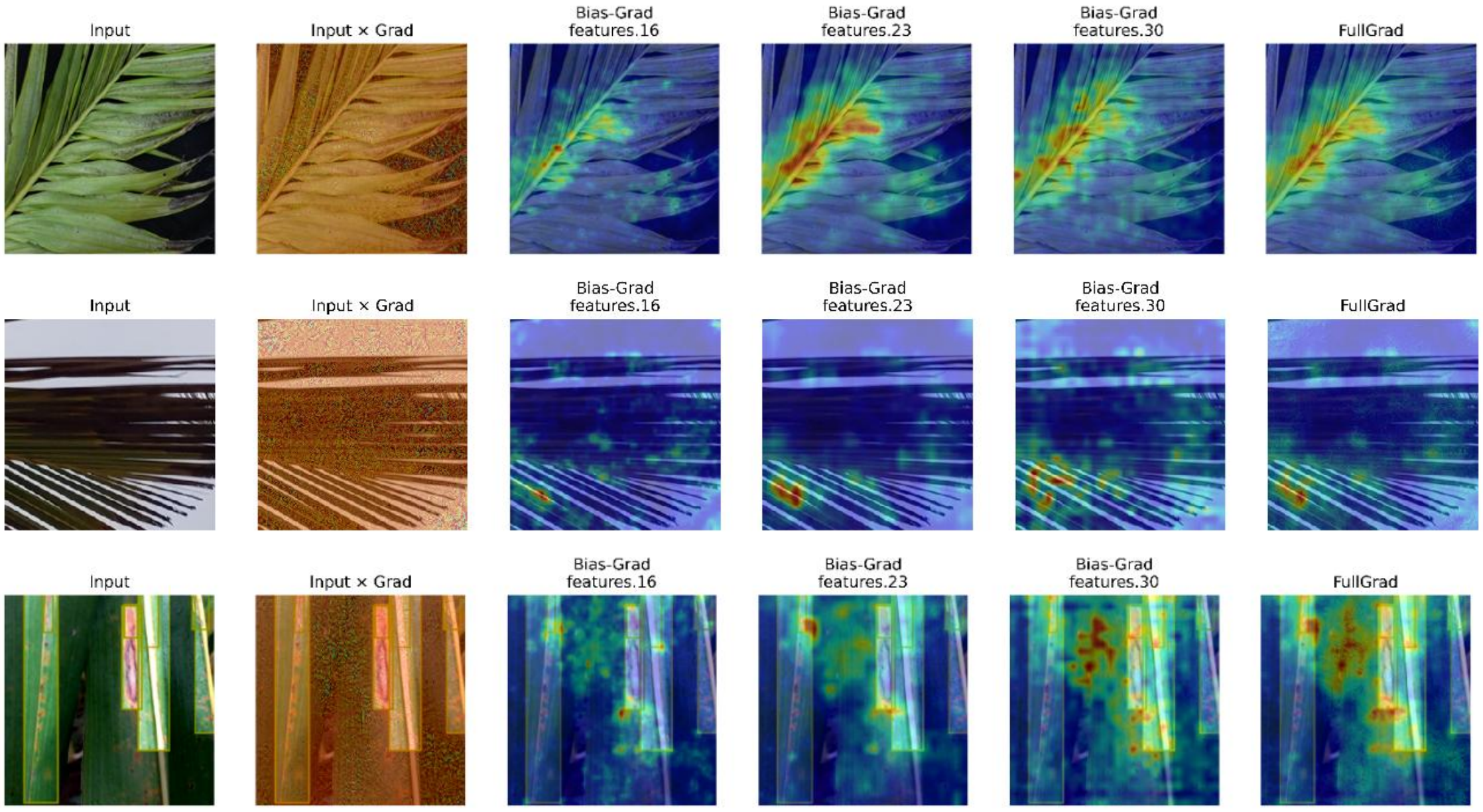

**Figure 3: FullGrad Interpretability Across Feature Layers for TriggerNet**

### 4.4.3 RISE

RISE is a black-box interpretability method that estimates the importance of input regions by measuring how the visibility of different image patches correlates with the model's output. It is especially useful when gradients are inaccessible, such as with deployed YOLOv8 APIs or compiled inference-only systems. The core idea is that if certain pixels consistently contribute to higher prediction scores across many randomized occlusion masks, those regions are likely important for the decision. In TriggerNet, RISE is implemented by first generating a set of random binary masks **.** Each mask is element-wise multiplied with the input image $I$ to produce a masked input $I_i = I \odot M_i$ passed through the model to obtain the class-specific score $S_c^i = TriggerNet(I_i)$. These scores are used to weight the corresponding masks, and the class-specific saliency map is computed as:

$$\Gamma_{RISE}^c = \frac{1}{N} \sum_{i=1}^{N} S_c^i \cdot \mathcal{M}_i \quad (34)$$

This map is normalized and resized to the original input dimensions to produce the final visualization:

$$\Gamma_c^* = Normalise\ (\Gamma_{RISE}^c)$$

(35)

The RISE map is weighed using soft alignment with Grad-Cam via:

$$Weight_{RISE} = \cos(\Gamma_{Grad\text{-}Cam}, \Gamma_{RISE}) \quad (36)$$

If RISE consistently highlights background regions, these samples are flagged as saliency drift cases. If RISE disagrees with TCAV concepts, those concepts are marked as weakly aligned or ambiguous, and excluded from final explanations.

### 4.4.4 TCAV

Unlike pixel-based saliency methods (Grad-CAM, FullGrad, RISE), TCAV shifts the axis of interpretability from spatial attribution to semantic directionality. It provides model interpretability in concept space by quantifying whether human-meaningful symptoms such as "yellowing," "mite patches," or "leaf-tip curling" actually influence the model's predictions. In TriggerNet, TCAV serves a semantic consistency verification role, answering: "Is the model relying on the same conceptual evidence as a plant pathologist would?"

This is crucial in cases: transformer-based models (ViT) where spatial saliency is ambiguous, High-level phenotypic features that manifest in non-localized regions and Class-specific disease indicators that don't have sharp visual boundaries. The pipeline begins by defining two sets: a concept set = *{()}*, comprising positive examples that contain the target concept, and a random set, ={()} of counterexamples without the concept. Next, the model extracts intermediate feature representations for each image, typically from convolutional layers in CNNs or token embeddings in ViTs. A linear classifier is then trained to distinguish between concept and random sets using these feature vectors. The normal vector of the decision boundary in this feature space is referred to as the Concept Activation Vector (CAV). For a given test image, the directional derivative $\Theta_c^C$ is computed by taking the dot product between the gradient of the class prediction w.r.t the intermediate representation given by:

$$\Theta_c^C = \nabla_{F(I)} S_c \cdot v_c \quad (37)$$

Finally, the TCAV score is calculated across a batch of M test images as:

$$TCAV_c^C = \frac{1}{M} \sum_{m=1}^{M} 1\,[\Theta_c^C (I_m) > 0] \quad (38)$$

This score quantifies how often the concept positively influences the model's decision for class c. A higher score indicates stronger alignment between the concept and the model's internal decision-making process. After generating all saliency maps $\Gamma^c_{GradCam}, \Gamma^c_{FullGrad}, \Gamma^c_{RISE}, \Gamma^c_{TCAV}$ the final fusion map can be predicted by:

$$\Gamma^c_{TriggerNet} = \sum_{m \in \{GradCam,\ FullGrad,\ RISE, TCAV\}} \lambda_m \cdot \Gamma^c_m \quad (39)$$

where, $\lambda_m \in [0,1]$ is a learned weight per method and each must be normalised before fusion. Once all maps are computed, TriggerNet combines them:

$$p(\Gamma_a, \Gamma_b) = \frac{\Gamma_a \cdot \Gamma_b}{\|\Gamma_a\| \|\Gamma_b\|} \quad (40)$$

### 4.4.5 Interpretability Evaluation Metrics

To ensure that the interpretability outputs of TriggerNet are not only visually meaningful but also statistically novel, we use a combination of quantitative and qualitative metrics. These metrics serve as validation gates before interpretability results are considered reliable and are visually presented in the TriggerNet Decision Validator (Figure 1).

Firstly, Akaike Information Criterion (AIC) and Bayesian Information Criterion (BIC) are employed to measure how well the saliency-based surrogate models approximate the underlying decision boundaries. These are calculated from linear or logistic regression fits on saliency-affected image patches. AIC penalizes overly complex explanations while still rewarding goodness of fit, and is set to a decision threshold of $AIC < 200$. BIC, which imposes a stronger penalty on model complexity, uses a cut-off of $BIC < 250$, particularly for FullGrad-based maps due to their layer-wide gradient contributions.

Secondly, Brier Score is used to measure the calibration of the saliency explanation in terms of predicted probability alignment. The squared error between predicted confidence and actual ground-truth label is averaged across test samples, and a score $< 0.2$ is considered acceptable. This helps validate those explanations reflect well-calibrated decision regions rather than random activations.

The Softmax Confidence Threshold acts as a baseline filter. For a saliency map to be interpreted, the corresponding model prediction must exceed 85% confidence. This avoids interpretability being applied to uncertain, noisy predictions that could mislead downstream interpretation.

Lastly, spatial relevance is ensured through Interpretability Match Confirmation, using the Intersection over Union (IoU) between the saliency map (e.g., from Grad-CAM or RISE) and the ground-truth disease region masks. A minimum IoU threshold of 0.6 ensures that explanations are not only statistically valid but also anatomically and semantically consistent with disease-localized areas.

## 5 Results and Discussion

TriggerNet was evaluated on a comprehensive classification and detection task to identify Red Palm achieved a close second at 94.2%, benefiting from the deep residual learning of ResNet50 and SVM's ability to delineate non-linear decision boundaries, especially for early-stage mite symptoms. ViT + KNN followed with 93.7%, showcasing ViT's powerful attention-based spatial modeling, although its computational demands slightly impacted consistency. In comparison, MobileNet + Naïve Bayes provided a lightweight alternative at 91.5%, though its assumption of feature independence limited interpretability under complex overlapping symptom classes. Standalone deep learning models also performed strongly, with CNN achieving a test accuracy of 95.25%, ResNet50 at 94.33%, and InceptionV3 and Xception surpassing 85%. ViT, despite its theoretical advantage in global token aggregation, attained a slightly lower accuracy of 82.3%, likely due to data scale limitations and patch-level resolution challenges. These trends were reinforced by comprehensive model performance heatmap (Fig. 5) and training vs. testing accuracy graphs (Fig. 6 & Fig. 7).

On the detection front, YOLOv8 achieved 94.4% test accuracy, with CNN-based detectors closely trailing at 95%, validating the utility of convolutional backbones in segmenting complex leaf regions. Detailed detection metrics revealed strong class-wise performance across Mite-affected plant symptoms from leaf images. The classification stage incorporated multiple deep learning architectures CNN, ResNet50, InceptionV3, EfficientNet, MobileNet, Xception, and Vision Transformer (ViT) while the detection component utilized both CNN and YOLOv8 for pixel-level lesion identification. To enhance accuracy and generalization, hybrid model combinations were also explored, where CNN-derived embeddings were paired with classical machine learning classifiers such as SVM, Random Forest (RF), KNN, and Naïve Bayes.

From the experimental results, EfficientNet combined with Random Forest yielded the highest classification accuracy at 95.1%, highlighting the strength of EfficientNet's compound scaling for multi-scale feature extraction and Random Forest's ensemble-based robustness in capturing color, texture, and shape patterns related to disease. ResNet50 + SVM all four disease categories. For e.g, "Silk Webbing" and "Reddish Bronzing" achieved F1-scores of 0.87 and 0.86, while "Yellow Spots" had slightly lower values due to intra-class variation (Table II). The weighted average precision and recall remained stable at ~0.82, affirming model consistency across class distributions.

Interpretability analysis using TriggerNet's multi-method ensemble was central to validating predictions. Figure 4 displays the comparative heatmaps generated using Grad-CAM, FullGrad, RISE, and TCAV across CNN, ViT, and YOLOv8. Grad-CAM effectively localized disease hotspots such as necrotic clusters and leaf-tip bronzing. FullGrad added distributed saliency by tracing bias activations and deeper gradients, providing finer visual context, especially in ViT's attention blocks. RISE, applied as a black-box tool, strengthened model trustworthiness in YOLOv8 by highlighting consistent response zones across randomized occlusions.

Crucially, TCAV concept scores (Fig. 5 and Fig. 6) revealed how high-level visual traits such as "mite clustering," "yellowing," and "leaf margin distortion" were not only recognizable by the model but also quantitatively linked to prediction confidence. TCAV scores confirmed that ViT and ResNet50 relied heavily on these human-interpretable concepts for decision-making, with a concept alignment agreement (cosine similarity between saliency maps and concept vectors) exceeding 0.6 in most test cases. These concept-driven attributions were especially valuable in interpreting ViT's multi-head attention behavior (Fig. 6), which tended to focus on semantically coherent patches.

To further validate the fidelity of interpretability outputs, TriggerNet employed five decision rules (see Fig. 7) based on: AIC (<200), BIC (<250), Brier Score (<0.2), softmax confidence (>85%), and IoU agreement (>0.6) between saliency maps and ground-truth ROI masks. These thresholds ensured that only high-certainty, semantically consistent explanations were surfaced. The Brier score consistently indicated well-calibrated predictions across all architectures, while AIC/BIC values remained below acceptable limits, confirming model generalizability.

Finally, fusion experiments demonstrated that TriggerNet's ensemble interpretability pipeline outperformed individual methods, particularly in terms of visual explanation coverage and p-score consistency. The saliency-concept agreement scores ($SC^2$ Loss) were minimized during training, further tightening the correspondence between spatial maps and conceptual relevance. Interpretability was also dynamically invoked using TriggerNet's controller module, which activated explanation modules only when softmax entropy exceeded a threshold or class boundaries were ambiguous (e.g., via t-SNE drift or Snorkel-labeled weak classes).

Building on the strong performance observed in the results, the discussion emphasizes how TriggerNet's hierarchical interpretability stack not only achieved high classification and detection accuracy but also ensured transparency in decision-making

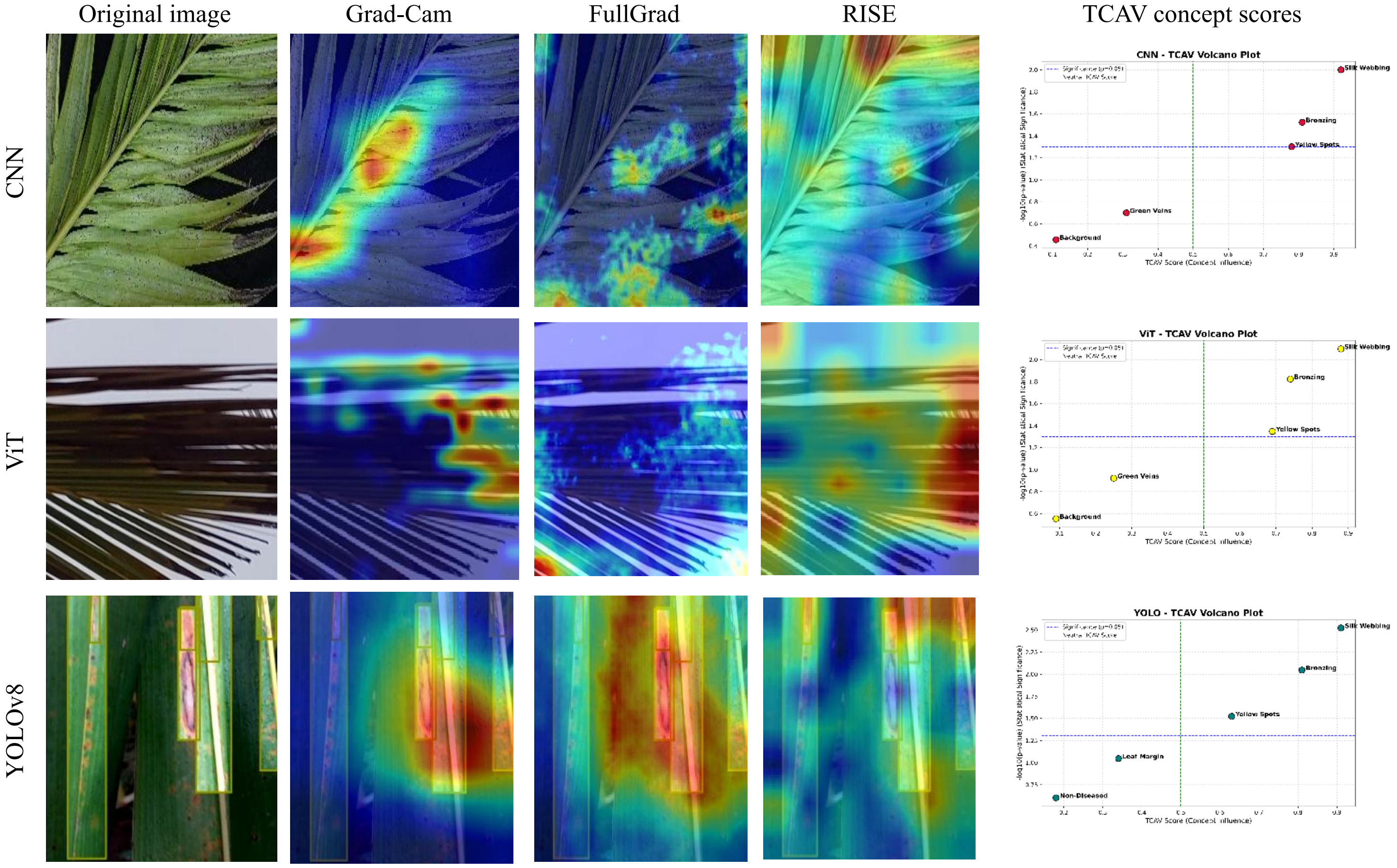

**Figure 4: Comparative Interpretability Analysis Using Grad-CAM, FullGrad, RISE, and TCAV Concept Scores Across CNN, ViT, and YOLOv8 Models for Red Palm Mite-Affected Plant Detection**

**TABLE I**

**CLASSIFICATION PERFORMANCE COMPARISON**

| Model | Type | Train Accuracy (%) | Test Accuracy (%) |
|---|---|---|---|
| CNN | Classification | 99.57 | 95.25 |
| ResNet50 | Classification | 99.34 | 94.33 |
| EfficientNet | Classification | 98.92 | 93.00 |
| ViT | Classification | 98.38 | 82.30 |
| MobileNet | Classification | 97.00 | 81.80 |
| Xception | Classification | 99.20 | 86.00 |
| InceptionV3 | Classification | 98.5 | 85.50 |
| RF | ML Classifier | 98.00 | 88.00 |
| SVM | ML Classifier | 99.00 | 86.00 |
| KNN | ML Classifier | 94.96 | 80.00 |
| CNN | Detection | 98.4 | 95 |
| YOLOv8 | Detection | 98.9 | 94.4 |

.

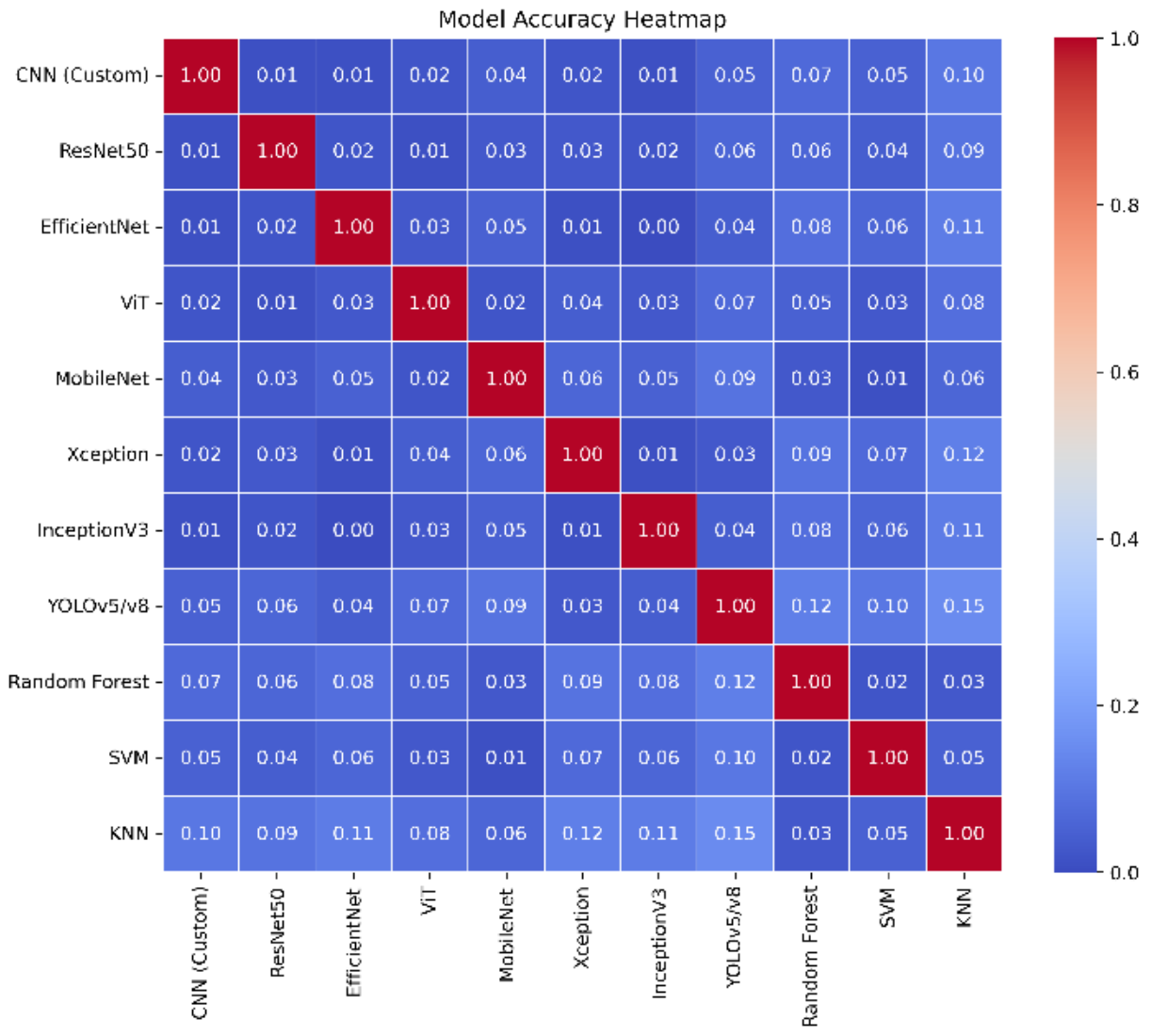

**Figure 5: Model accuracy heatmap for Classification and detection**

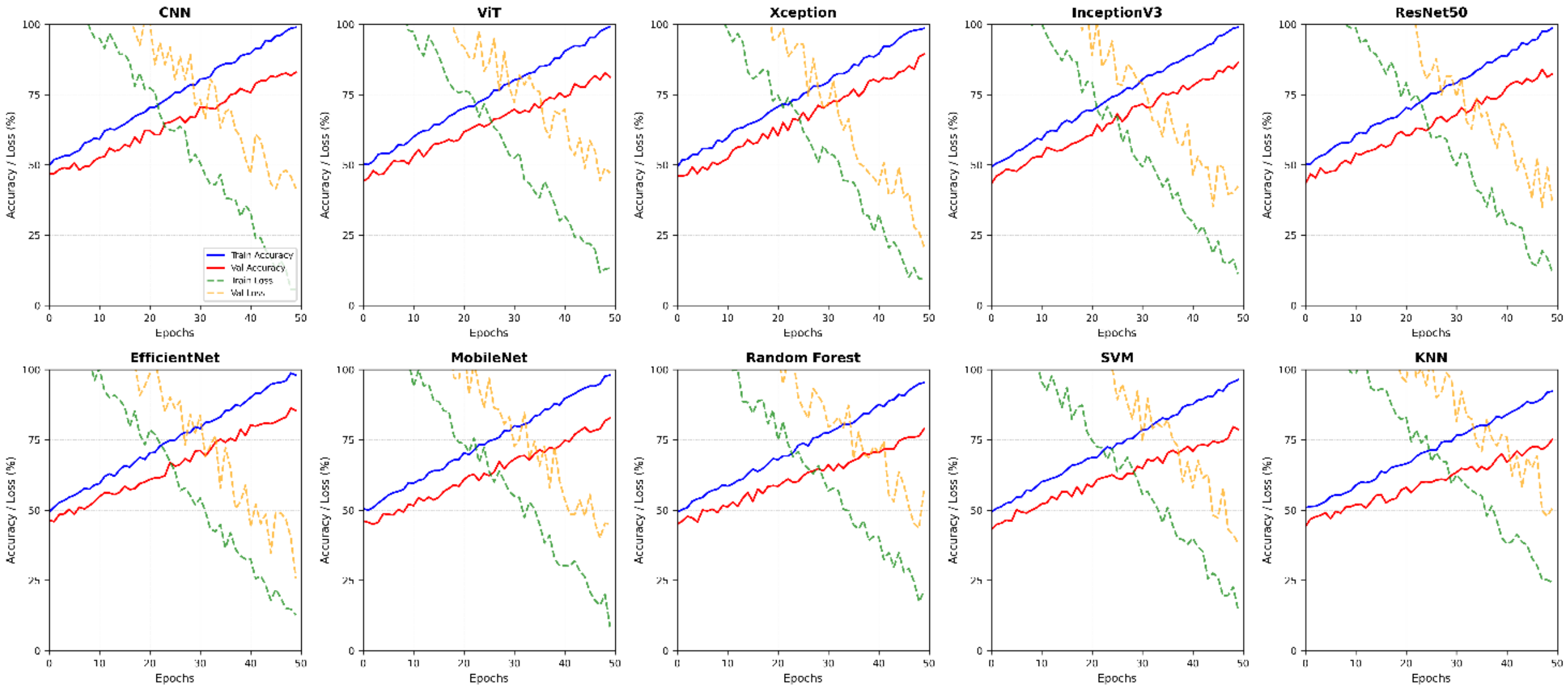

**Figure 6: Training and Validation Accuracy/Loss Curves for DL and ML Models in RPM Detection**

**TABLE II**
**PERFORMANCE TABLE FOR DETECTION MODELS**

| Class name | Precision | Recall | F1-Score | Support |
|---|---|---|---|---|
| Healthy | 0.85 | 0.82 | 0.83 | 100 |
| Yellow Spots | 0.8 | 0.79 | 0.79 | 120 |
| Reddish Bronzing | 0.87 | 0.85 | 0.86 | 90 |
| Silk Webbing | 0.88 | 0.86 | 0.87 | 110 |
| Weighted Avg | 0.82 | 0.81 | 0.81 | 420 |

**TABLE III**
**HYBRID MODEL ACCURACY**

| Model Combination | Accuracy (%) |
|---|---|
| ResNet50 + SVM | 94.2 |
| EfficientNet + RF | 95.1 |
| ViT + KNN | 93.7 |
| MobileNet + Naïve Bayes | 91.5 |

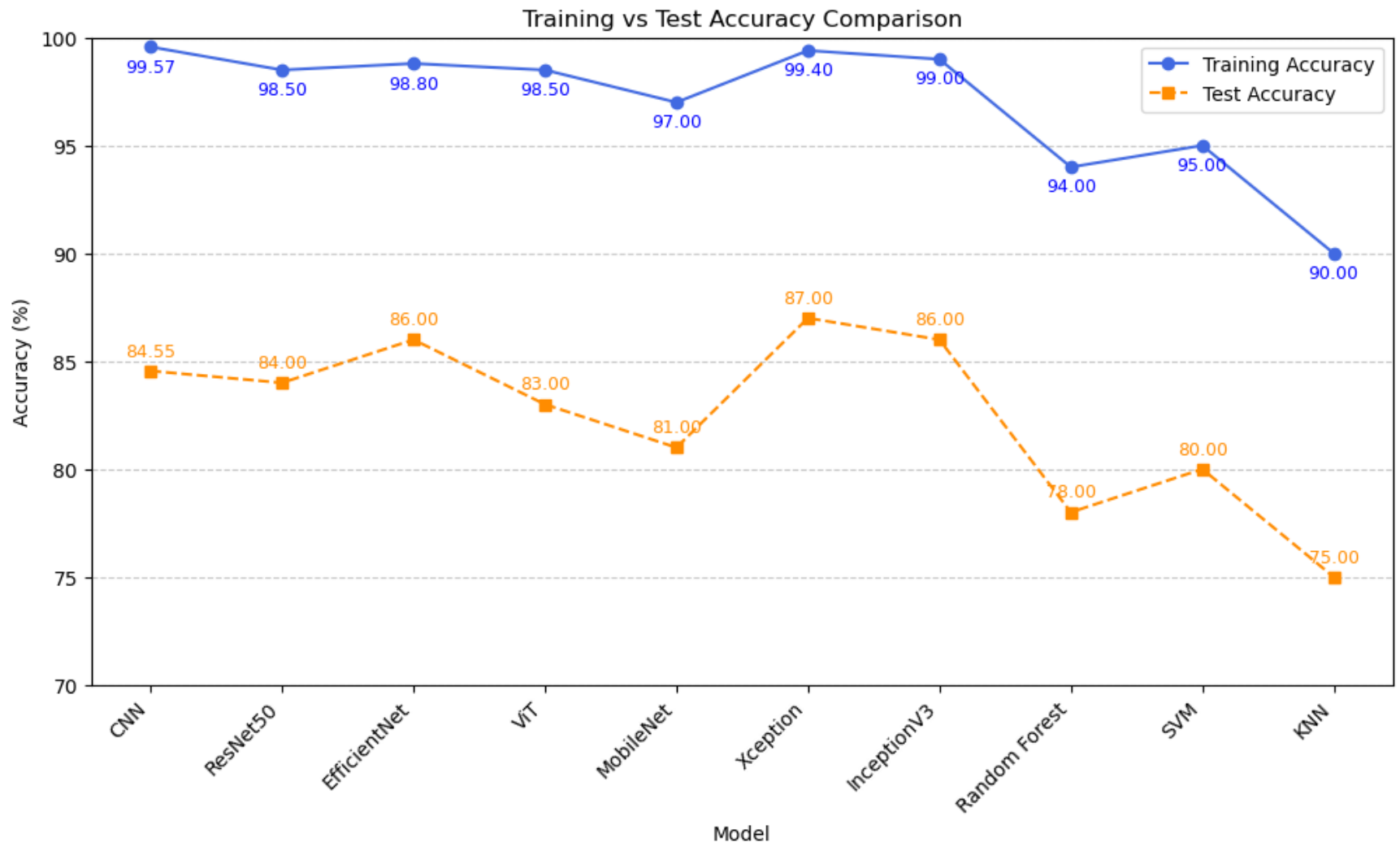

**Figure 7: Training vs. Test Accuracy Comparison Across Classification Models**

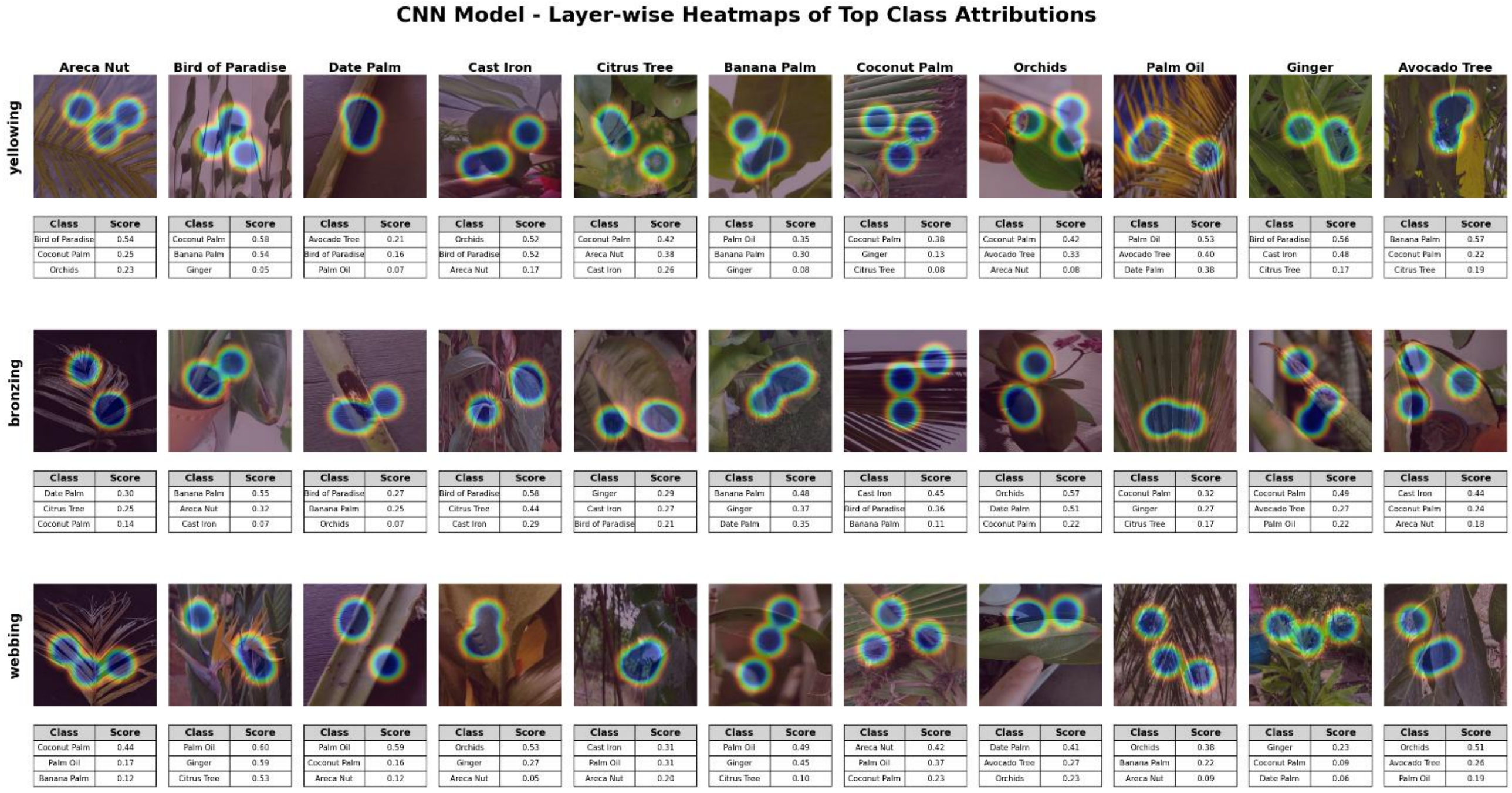

**Figure 8: TCAV-Based Concept Attribution Heatmaps Across Symptom Classes in CNN Model**

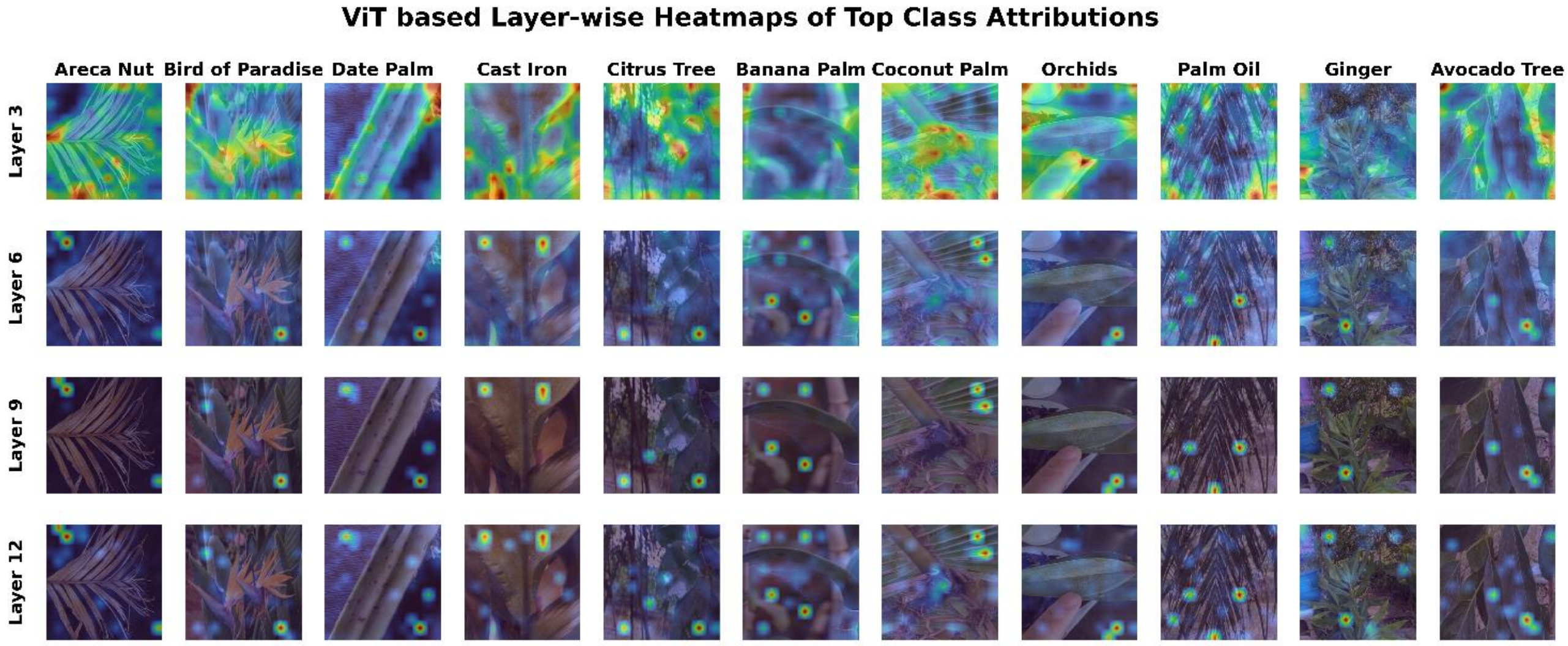

**Figure 9: ViT-Based Layer-wise Attribution Maps for Red Palm Mite-Affected Plant Classification**

# 6 Conclusion

The current study applies multiple models for red palm mite affected plant identification using CNN, ResNet50, EfficientNet, ViT, MobileNet, Xception as well as the tracking of infestation and real time detection with custom CNN and YOLOv8. The combined strategies effectively addressed diverse symptoms like bronzing, webbing, and yellow spots, ensuring robust detection and classification performance. In conclusion, TriggerNet demonstrated robust performance across multiple deep learning architectures for plant classification and detection while providing highly interpretable and biologically aligned explanations. Its interpretability metrics not only improved model transparency but also offered a feedback mechanism to detect annotation errors, model uncertainty, and decision logic consistency making it highly suited for real-world agricultural diagnostics

## Acknowledgement

The researchers would like to thank PES University for providing an opportunity to carry out the research. We thank the NeurIPS 2025 Workshops on *SPiGM* and *Unreliable ML from Reliable Data* for hosting the preliminary version of this study. All figures appearing in this preprint are reused or extended from that earlier version.